%% file: main.tex
\documentclass[10pt,twocolumn,letterpaper]{article}

\usepackage{cvpr}

\usepackage{graphicx}
\usepackage{makecell}
\usepackage{tikz}
\usetikzlibrary{positioning}
\usepackage{xcolor}
\usepackage{array}

\usepackage[normalem]{ulem}


\definecolor{cvprblue}{rgb}{0.21,0.49,0.74}
\usepackage[pagebackref,breaklinks,colorlinks,allcolors=cvprblue]{hyperref}

\def\paperID{*****} % *** Enter the Paper ID here
\def\confName{CVPR}
\def\confYear{2025}

\title{Deep Learning-based Animal Behavior Analysis: Insights from Mouse Chronic Pain Models}

\author{
Yu-Hsi Chen\\
Institute of Information Science (IIS)\\
Academia Sinica, Taipei, Taiwan\\
{\tt\small franktpmvu@iis.sinica.edu.tw}
\and
Wei-Hsin Chen\textsuperscript{\textdagger}\\
Institute of Biomedical Sciences (IBMS)\\
Academia Sinica, Taipei, Taiwan\\
{\tt\small vic7538@ibms.sinica.edu.tw}
\and
Chien-Yao Wang*\\
Institute of Information Science (IIS)\\
Academia Sinica, Taipei, Taiwan\\
{\tt\small kinyiu@iis.sinica.edu.tw}
\and
Hong-Yuan Mark Liao*\\
Institute of Information Science (IIS)\\
Academia Sinica, Taipei, Taiwan\\
{\tt\small liao@iis.sinica.edu.tw}
\and
James C. Liao\\
Institute of Biological Chemistry(IBC)\\
Academia Sinica, Taipei, Taiwan\\
{\tt\small liaoj@gate.sinica.edu.tw}
\and
Chien-Chang Chen\textsuperscript{\textdagger}\\
Institute of Biomedical Sciences (IBMS)\\
Academia Sinica, Taipei, Taiwan\\
{\tt\small ccchen@ibms.sinica.edu.tw}
}

\begin{document}

\providecommand{\needtoupdate}[1]{\textcolor{red}{needtoupdate}}

\maketitle

\renewcommand{\thefootnote}{\fnsymbol{footnote}}
\footnotetext[1]{Corresponding authors for Computer Vision:
\texttt{kinyiu@iis.sinica.edu.tw},
\texttt{liao@iis.sinica.edu.tw}}
\footnotetext[2]{Corresponding authors for Biomedicine:
\texttt{vic7538@ibms.sinica.edu.tw},
\texttt{ccchen@ibms.sinica.edu.tw}}
\renewcommand{\thefootnote}{\arabic{footnote}}

\input{sec/0_abstract_mm_eng}

\input{sec/1_intro_multi_media_eng}

\input{sec/2_survey_mm_eng}

\input{sec/3_dataset_mm_eng}

\input{sec/4_method_mm_eng}
\input{sec/6_ana_dis_mm_eng}
\input{sec/7_con_mm_eng}
{
    \small
    \bibliographystyle{ieeenat_fullname}
    \bibliography{main}
}
\end{document}

%% file: sec/0_abstract_mm_eng.tex
\begin{abstract}

Assessing chronic pain behavior in mice is critical for preclinical studies.  However, existing methods mostly rely on manual labeling of behavioral features, and humans lack a clear understanding of which behaviors best represent chronic pain.  For this reason, existing methods struggle to accurately capture the insidious and persistent behavioral changes in chronic pain.
This study proposes a framework to automatically discover features related to chronic pain without relying on human-defined action labels.  Our method uses universal action space projector to automatically extract mouse action features, and avoids the potential bias of human labeling by retaining the rich behavioral information in the original video.
In this paper, we also collected a mouse pain behavior dataset that captures the disease progression of both neuropathic and inflammatory pain across multiple time points.
Our method achieves 48.41\% accuracy in a 15-class pain classification task,
significantly outperforming human experts (21.33\%) and the widely used method B-SOiD (30.52\%).
Furthermore, when the classification is simplified to only three categories, i.e., neuropathic pain,
inflammatory pain,
and no pain, then our method achieves an accuracy of 73.1\%, which is notably higher than that of human experts (48\%) and B-SOiD (58.43\%).
Finally, our method revealed differences in drug efficacy for different types of pain on zero-shot Gabapentin drug testing, and the results were consistent with past drug efficacy literature.  This study demonstrates the potential clinical application of our method, which can provide new insights into pain research and related drug development.

\providecommand{\keywords}[1]{\textbf{\textit{Index terms---}} #1}

\keywords Chronic Pain, Behavior Analysis, Deep Learning, Universal Action Space, Neuropathic pain, Inflammatory pain.

\end{abstract}

%% file: sec/1_intro_multi_media_eng.tex
\section{Introduction}
\label{sec:intro}

\begin{table*}[h]
    \centering
    \caption{Comparison of Different Approaches in Pain Analysis}
    \label{tab:pain_analysis_comparison}
    \vspace{-8pt}
    \resizebox{1.0\textwidth}{!}{ % Adjust table size for two-column format
    \begin{tabular}{|l|c|c|c|c|c|c|}
    \hline
    \textbf{Approach} & 
    \parbox{1.6cm}{\textbf{Reflex} \\ \textbf{Response}}\rule{0pt}{4ex} & 
    \parbox{2.5cm}{\textbf{Acute/Chronic} \\ \textbf{Pain}}\rule[-3ex]{0pt}{4ex} &
    \textbf{Pain Cause} &
    \textbf{Human Operation} &
    \textbf{Human Labeling} &
    \textbf{Human Judgment} \\ 
    \hline
    Hargreaves Test         & \checkmark & x & x & 20 min per mouse & Observation only & \checkmark \\ 
    \hline
    Von Frey Test     & \checkmark & x & x & 20 min per mouse & Observation only & \checkmark \\ 
    \hline
    Gait Analysis\cite{cat_walk_Xu_2019} & x & \checkmark & x & 10 min/day pre-training for 5 days
 & Observation only & \checkmark \\ 
    \hline
    \parbox{4cm}{Home Cage Monitoring \\ \cite{homecagescan_roughan2009automated,LABORAS_Castagn__2012}} \rule{0pt}{3.7ex}
    &  & \checkmark & \checkmark & Only place mice in box & Requires manual feature selection
 & \checkmark \\ 
    \hline
    \parbox{4cm}{Other ML-Based Methods \\ ~\cite{deep_lab_cut_mathis2018deeplabcut,b_soid_Hsu_2021,simBA_Goodwin2024,asoid_tillmann2024soid}}\rule{0pt}{3.7ex}
    &  & \checkmark & \checkmark & Only place mice in box & Requires manual feature selection
    
 & \checkmark \\ 
    \hline
    \textbf{Ours}         &  & \checkmark & \checkmark & Only place mice in box & x & x \\ 
    \hline
    \end{tabular}
    }
    \vspace{-18pt}
\end{table*}

Using mice for preclinical research is a critical step in drug development,
and assessing the pain response in mice enables researchers to explore the underlying mechanisms of pain and evaluate the eﬀicacy of analgesic treatments.
Mouse models are commonly used to evaluate the efficacy and safety of new treatments before entering human clinical trials~\cite{safety_Gutmann_2006}.
In 2009, Mogil~\cite{summarize_Mogil_2009} emphasized the limitations of existing mouse models in reflecting the complexity of human chronic pain,
and called for refined approaches that better capture spontaneous behaviors and psychological comorbidities.
Mouse models remain essential for investigating pain mechanisms and evaluating the efficacy of potential analgesics.
Several studies have used mouse models to test analgesic effects of various natural substances such
as coconut water~\cite{yong_coconut_dini2021analgesic} and cactus flowers~\cite{Opuntia_ficus-indica_flowers_layachi2023assessment},
demonstrating the model's versatility in translational research.
Recently, Barcelon et al.~\cite{human_and_mice_pain_barcelon2023sexual} revealed the mechanisms underlying central sensitization in chronic pain and highlighted significant sex differences involving immune and glial responses,
including neural sensitization, inflammation, and neuropathic processes.
These physiological phenomena are important causes of chronic pain in humans.
While such studies highlight the utility of mouse models,
the assessment of pain often relies on behavioral observations that are subtle, subjective, and labor-intensive, which poses significant challenges to scalability and reproducibility.

Methods for analyzing pain in the literature are mainly divided into three categories, namely reflex testing, autonomous behavior analysis, and reward and punishment testing~\cite{summarize_Mogil_2009,self_Autonomous_behavior_analysis_Tappe_Theodor_2014}.  
Reflex testing mainly evaluates pain sensitivity by measuring the reaction time of mice to external stimuli, such as Hargreaves test, Hot Plate Test, and Von Frey Test.  
As for autonomous behavior analysis, pain is analyzed by monitoring and counting the natural behavioral effects of mice, such as movement, posture, and social interaction.  
Methods for natural behavior analysis include DigiGait~\cite{cat_walk_Xu_2019} and CatWalk~\cite{cat_walk_Xu_2019}, which use infrared rays for gait analysis, as well as Home CageScan~\cite{homecagescan_roughan2009automated} and LABORAS~\cite{LABORAS_Castagn__2012}, which observe behavioral changes in mice for a long time. 
Reward-and-punishment-related tests use differential giving mechanisms to stimulate the learning, memory, and decision-making processes of mice.  
For example, conditioned place preference (CPP)~\cite{cpp_tzschentke2007review} uses a special place design to evaluate the behavioral response of mice to pain relief.

In fact, all the above methods have clinical limitations or practical difficulties.  
For example, reflex testing only captures spinal reflexes, and if it were to be widely used in standardized acute pain research, it would not fully capture the persistence and emotional components of chronic pain.  
Current autonomous behavior analysis methods mostly rely on expensive equipment and require analysis based on behavior categories defined by human experts.  
As for the reward-and-punishment test, it is easily affected by the mouse's ability to learn the environment, so it is difficult to provide a correlation analysis for objective pain behavior assessment.  
In addition, these methods usually rely on the intervention of human experts and are therefore prone to observer bias~\cite{Observer_bias_TUYTTENS2014273}, thus limiting the objectivity and reproducibility of the data.

In recent years, there has been tremendous progress in research on the use of machine learning for pain behavior analysis~\cite{deep_lab_cut_mathis2018deeplabcut,mars_Segalin2020.07.26.222299,b_soid_Hsu_2021,asoid_tillmann2024soid,simBA_Goodwin2024}.  
It minimizes observer bias with the advantage of minimizing human intervention, and at the same time provides a detailed and efficient behavioral pattern analysis tool.  
In 2018, Mathis et al.~\cite{deep_lab_cut_mathis2018deeplabcut} proposed DeepLabCut, which is a transfer learning technology.  DeepLabCut combines the DeeperCut model trained in human pose detection~\cite{deepercut_insafutdinov2016deepercut} with a small amount of annotated data to quickly adapt and accurately track the body parts of any experimental animal, achieving excellent results.  
Due to its high performance and accuracy, DeepLabCut has become a powerful tool for many animal behavior analysis and biomedical research.  
It has been widely used in fields such as movement analysis, pain behavior observation, and neuroscience experiments, and has become one of the standard pre-processing methods for animal experiments.  
In 2021, Hsu et al.~\cite{b_soid_Hsu_2021} developed the unsupervised learning method B-SOiD. They automatically clustered behavioral patterns using the mouse's limb trajectories, and they also found that this method could detect behaviors that humans could not recognize.  
In 2024, Goodwin et al.~\cite{simBA_Goodwin2024} developed the supervised learning method SimBA.  They invited researchers to manually label areas of interest and developed a system that can automatically analyze animal limbs and trajectories in the area, hoping to discover different behavioral patterns. 
However, most existing methods still rely on action categories defined by human experts to establish correlation with pain behaviors, making it difficult to directly identify pain behaviors.

In order to solve the problems of existing methods,
we develop a very general platform that can interpret and analyze various animal behaviors.
In addition to using this platform to analyze chronic pain in mice,
it can also be extended to other animal and human action-related analysis in the future.
To develop the above platform,
we used the existing large-scale human behavior dataset,
Kinetics-600,
as the basis to train a Universal Action Space (UAS),
and this UAS can be used to represent a wide variety of complex actions. 
The key concept we rely on is that human behaviors are composed of very complex combinations of actions,
and when a large and diverse enough human behavior dataset is used to construct the UAS,
the space it unfolds must be a superset of the action spaces of other animals.  In other words, the action space unfolded by various actions of general animals can be regarded as a subspace of the UAS unfolded by complex human behaviors.

In this paper, we propose to directly learn pain classification through long-term videos of natural mouse activities, which does not require any external stimulation done by human beings.  
As shown in Table~\ref{tab:pain_analysis_comparison}, our proposed method can not only identify acute pain, but can even further analyze the causes of pain, such as neuropathic and inflammatory pain.  
The main contributions of this study are summarized as follows:
\begin{itemize}
    \item A fully end-to-end mouse autonomous behavioral pain analysis system is proposed, which can analyze pain types and pain causes simultaneously.
    \item Use the existing large-scale human action dataset Kinetics-600 to train a UAS which can be used to represent actions, and experimentally demonstrate that this UAS can include most of the action space projected by mouse actions.
    \item Construct a video dataset of mouse pain behavior covering long-term and diverse time stamps,
    including two pain types (acute and chronic) and a painless control group, with three etiologies (nerve injury, inflammation, and no injury), 
    for a total of 632 videos.
    \item Conduct comprehensive benchmark testing and real-world scenario evaluation, and it is confirmed that the performance of the proposed method is significantly better than the commonly used B-SOiD analysis method and human expert judgment.
\end{itemize}

%% file: sec/2_survey_mm_eng.tex
\section{Related Works}
\label{sec:survey}

\subsection{Behavioral Analysis of Pain in Mice}
\label{subsec:Mice pain}

Existing research assesses chronic pain by analyzing short-term action statistics.  
These methods all follow a logical assumption: By observing movement frequency, movement duration, and movement changes, researchers can infer pain-related behaviors~\cite{homecagescan_dance_doi:10.1073/pnas.0610779104_2007,homecagescan_roughan2009automated,summarize_Mogil_2009,LABORAS_Castagn__2012,homecagescan_day_night_ADAMAHBIASSI2013306_2013,machine_learning_auto_ranking_Wotton_2020,b_soid_Hsu_2021,homecagescan_ongoing_pain_BURANDJR2023100113_2023,asoid_tillmann2024soid}.  
Traditionally these statistics have been collected and analyzed manually, requiring researchers to define key pain-related actions and quantify these actions.  
With the advancement of computer vision and machine learning, researchers have developed machine learning-based mouse behavior recognition systems to further improve the accuracy of pain assessment. 
Mouse Action Recognition System (MARS)~\cite{mars_Segalin2020.07.26.222299} used pose estimation and deep learning to classify mouse social behaviors such as aggression, mating, and exploration.  
Simple Behavior Analysis (SimBA)~\cite{simBA_Goodwin2024} manually marks area of interest and classifies actions based on the relative positions of key points.

The above methods improve the accuracy of behavior identification, but relying solely on human-annotated label data will suffer from observer bias and limited category coverage~\cite{Observer_bias_TUYTTENS2014273}.  
To reduce the cost of manual annotation and lower down the human bias, Hsu et al.~\cite{b_soid_Hsu_2021} developed the unsupervised learning method B-SOiD.  
This method can automatically classify behavioral patterns, and its most special feature is that it can discover behaviors that were not discovered by human intervention in the process.  
However, the above approach also poses a challenge, that is, rare but important key behaviors may be ignored if they do not occur frequently.  
In order to seek a balance between accuracy and discoverability, Hsu et al.~\cite{asoid_tillmann2024soid} developed A-SOiD based on B-SOid.  They use only small amounts of human-labeled data, allowing algorithms to autonomously explore new categories of behavior.  
This method minimizes observer bias while having the ability to detect rare 
behaviors, making it an excellent solution for chronic pain research.  
However, these unsupervised or semi-supervised methods still face the difficulty of selecting key points.  
If researchers ignore important indicators of pain (such as whisker contraction, paw flexion, and subtle claw protective behaviors~\cite{pain_face_Fried_2020}), machine learning methods will not be able to learn these features based on key points, thus limiting the ability to autonomously discover new behaviors.

\subsection{From Short-Term Action Recognition to Long-Term Behavioral Analysis in Deep Learning}
\label{subsec:Action recognition}
The purpose of our research is to use machine learning to learn the behaviors of various animals, and then to perform various applications by directly comparing unknown behaviors with previously learned behaviors. In the past few years, some methods have been proposed to describe and identify behaviors using CNN\cite{c3d_tran2015learningspatiotemporalfeatures3d}, CNN+LSTM~\cite{cnn_lstm_7178838}, or Transformer~\cite{attention_is_all_you_need_vaswani2023attentionneed}. C3D is proposed by Tran et al. ~\cite{c3d_tran2015learningspatiotemporalfeatures3d}, it uses 3D convolution on the frames of a video, trying to extend the traditional CNN to the time domain, thereby effectively capturing the short-term spatio-temporal features.  
However, the C3D model is difficult to handle long-term dependencies because it can only be used for fixed time windows, thus limiting its ability to be used to analyze long-term behavior.  
On the other hand, Sainath et al.~\cite{cnn_lstm_7178838} combined CNN with LSTM to handle temporal dependencies, but there are still many shortcomings in handling very long-term dependencies~\cite{do_rnn_have_long_memory_zhao2020rnnlstmlongmemory}.  
With the development of the self/cross attention mechanism, Transformer-based models such as ViVit~\cite{vivit_arnab2021vivitvideovisiontransformer} and TimeSFormer~\cite{timesformer_bertasius2021spacetimeattentionneedvideo} have been applied in video action recognition. Different from the CNN-LSTM architecture, Transformer can capture global dependencies on the timeline, thereby achieving a more comprehensive action representation capability.  
However, a major limitation of the Transformer-based architecture is that its time complexity grows with the square of the time series length~\cite{attention_is_all_you_need_vaswani2023attentionneed}, which creates a significant computational burden.

Unlike Transformer, the time complexity of S4~\cite{s4_gu2022efficientlymodelinglongsequences} grows linearly with the length of the time series.  
The main reason is that the S4 model does not rely on self-attention mechanism, but uses structured state space modeling to capture
long-term dependencies, and thus can more accurately track long-term behavioral changes.  
Recently some models that combine Transformer's short-term action
features with long-term modeling of the S4 model have been proposed, such as ViS4mer~\cite{vis4mer_islam2023longmovieclipclassification}.  
But how to effectively integrate short-term action feature extractor with long-term behavior classifier is still an issue that needs to be solved.
Although pre-trained short-term action feature extractor provides rich semantic information, they also include movements that are not relevant to behavioral classifications of chronic pain.  
In this paper, we combined the time-series based mask mechanism on the basis of ViS4mer, 
which can selectively reduce the impact of irrelevant motions to improve the ability to identify chronic pain.

%% file: sec/3_dataset_mm_eng.tex
\section{Dataset Construction, Software Systems for Training and Analysis}
\label{sec:dataset_construction}

In this section, we will introduce how to set up the experimental scenario, how to collect data, how to use the existing human behavior dataset kinetics-600 to build a UAS, and how to build a subspace on top of the UAS that can analyze mouse pain.

\subsection{Experiment Site Design}

\begin{figure}
    \centering
    \includegraphics[width=1\linewidth]{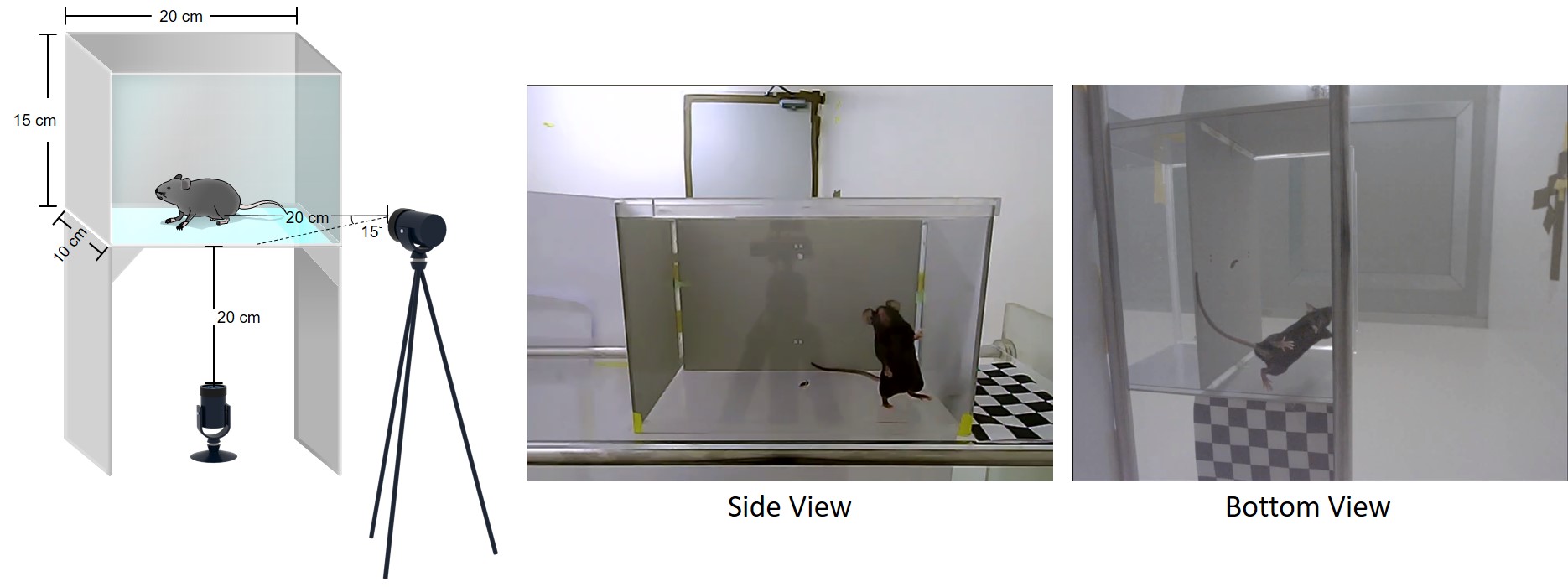}
    \caption{Visualization of how data are collected: Some videos contain both side-view and bottom-view perspectives.}
    \label{fig:dataset_visualize}
\end{figure}

The experiment site design is shown in Figure~\ref{fig:dataset_visualize}.  
First, we designed a cage for mice to move around.  This cage consists of three white sides and three transparent sides.  
Among them, the transparent plate on the side is used to observe the movements and behaviors of the mouse, the transparent plate below is used to observe the gait and reaction of the mouse, and the transparent plate above is used to observe the trajectory of the mouse.  
In order to allow the mice enough space to move around, the length, width, and height of this cage are 20cm $\times$ 10cm $\times$ 15cm.  
In order to reduce the burden on the mice, there are no decorations or patterns in the cage.  
In each cage, we used two cameras (Basler acA1300-200uc and a USB webcam) to shoot side-view and bottom-view videos respectively. 
The camera and the cage are placed vertically, with a distance of 40 cm.  
The mice were placed in their cages for each data collection session, and five minutes of video of the mice's natural activities were taken.

\subsection{Data Collection Settings}
\label{sec:data_collection_settings}

In the Mouse Pain Analysis Dataset, we collect videos of mice with different pain types and causes.  
All protocols were approved by the Institutional Animal Care \& Use Committee and followed the standards of the National Institutes of Health and the International Association for the Study of Pain Guidelines for the Use and Care of Laboratory Animals.  
We divided age-matched mice of 8 to 12 weeks and housed them in an environment with constant temperature (26°C), humidity (65\%), and light/dark cycle (7:00-19:00 illumination), and have free access to food and water.  
All animals were monitored daily throughout the study to ensure their overall good health.  In all experiments, we used C57BL/6J NarL mice.  
Regarding the causes of pain, we recorded videos of inflammatory pain, neuralgia, and healthy mice.
\begin{itemize}
    \item Inflammation model (formalin): 10~$\mu L$ of 5\% formalin in phosphate-buffered saline (PBS) was injected subcutaneously into the plantar surface of the left hind paw.
    \item Neuralgia model (spared nerve injury, SNI): The left common peroneal and sural nerves were dissected, and the tibial nerve was left intact~\cite{SNI_PMID:31009420}.
    \item Healthy mice: The mice have not received any surgery or drug interference and are the reference benchmark for normal behavior.
\end{itemize}

In terms of pain types, we recorded the pain behavior of mice for up to 21 days, as shown in Table~\ref{tab:dataset}, and sampled at the following time points:
\begin{itemize}
    \item D0: The initial state of the mice was recorded.  At this time, the mice were all healthy. 
    \item 1 min: The acute sustained pain response in mice induced by formalin injection was recorded.
    \item 2h: The sustained pain response in mice induced by formalin injection was recorded.
    \item D3: Early adaptation to pain in formalin-injected and neurosurgery mice was recorded.
    \item D5, D7, D14, D21: Tracking medium- and long-term chronic pain responses in mice.
\end{itemize}
In this dataset,
we collected 255 videos of healthy mice (including D0 mice),
153 videos of neuralgia mice,
and 224 videos of inflammatory pain mice,
for a total of 632 videos (five minutes each).  Since sampling time points correspond directly to pain types, no additional manual annotation is required.

\begin{table}[h]
    \centering
    \caption{Dataset Distribution Across Pain Conditions}
    \label{tab:dataset}
    \resizebox{0.3\textwidth}{!}{ % Adjust table size for two-column format
    \begin{tabular}{|l|c|c|c|}
    \hline
    \textbf{Time} & \multicolumn{3}{c|}{\textbf{Number of Videos}} \\ \hline
    
    &\textbf{Formalin} & \textbf{SNI} & \textbf{Control} \\ \hline
    \textbf{D0 } & \multicolumn{3}{c|}{147} \\
    \hline
    1 min  & 56  & -  & -  \\ 
    2h     & 56 & -  &  - \\ 
    D3     & 28 & 60  & 30  \\ 
    D5     & 28 & -  &  - \\ 
    D7     & 28 & 31  & 30  \\ 
    D14    & 28 & 31  & 30  \\ 
    D21    & - & 31  & 18  \\ \hline
    \textbf{Total} & 224 & 153 & 108 \\ \hline
    \end{tabular}
    }
\end{table}

\subsection{Software Systems for Training and Analysis}

In this work, we adopted Video Swin Transformer (VST)~\cite{video_swin_transformer_liu2021videoswintransformer} to extract spatial-temporal action features from training videos to train a UAS.
Then ViS4mer~\cite{vis4mer_islam2023longmovieclipclassification} is adopted to construct a pain type analyzer. We follow the preset hyperparameters of the above model to train or fine-tune the model.  In pain type analyzer, we additionally add focal loss~\cite{focal_loss_lin2018focallossdenseobject} for training.  The main training hyperparameters include: learning rate 0.001, weight decay 0.01, batch size 16, optimizer Adam, hidden dimension 1024, dropout 0.2. If the performance does not improve after training for 10 epochs, learning rate will be adjusted to 0.2 times, and the model will be trained with NVIDIA 1080 Ti for 200 epochs.

\subsection{Construction of UAS}
\label{subsec:construction of UAS}

\begin{figure}
    \centering
    \includegraphics[width=1\linewidth]{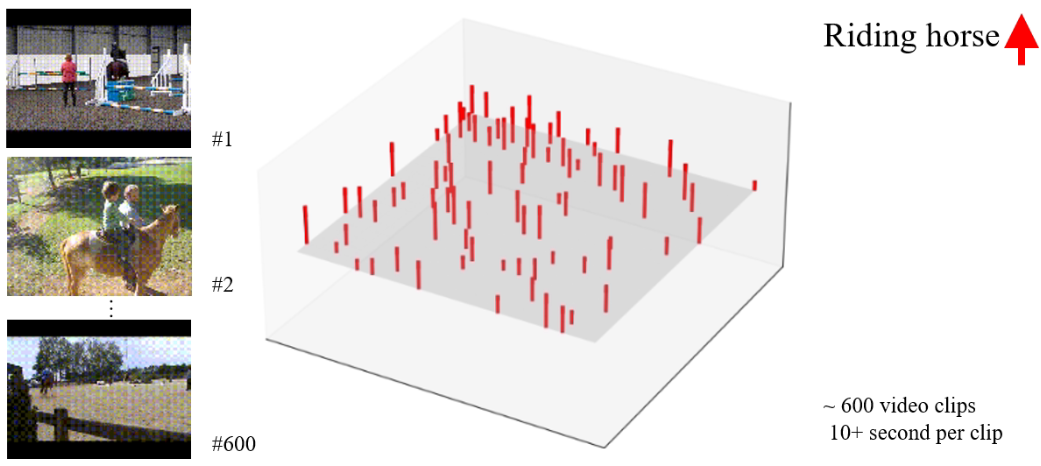}
    \caption{Riding horse.}
    \label{fig:riding_horse}
\end{figure}

\begin{figure}
    \centering
    \includegraphics[width=1\linewidth]{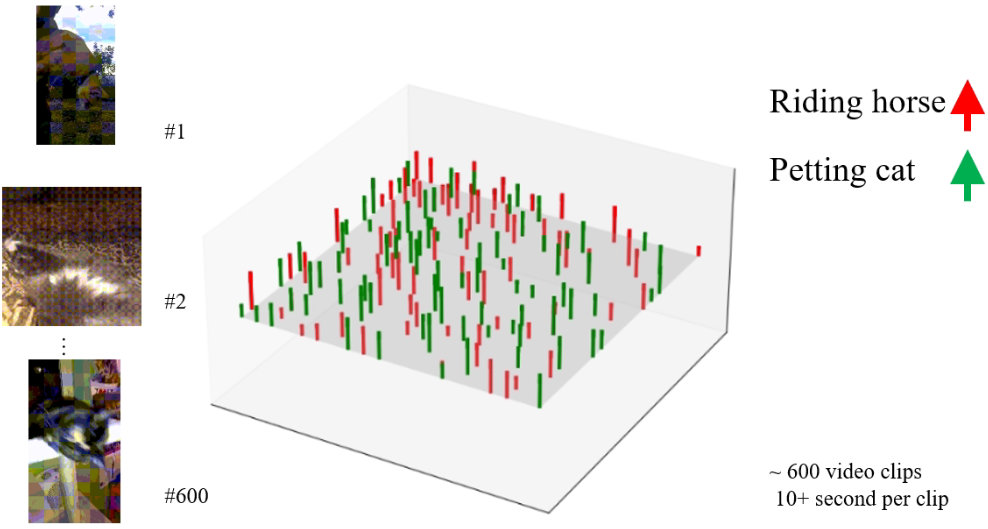}
    \caption{Petting cat.}
    \label{fig:petting_cat}
\end{figure}

\begin{figure}
    \centering
    \includegraphics[width=1\linewidth]{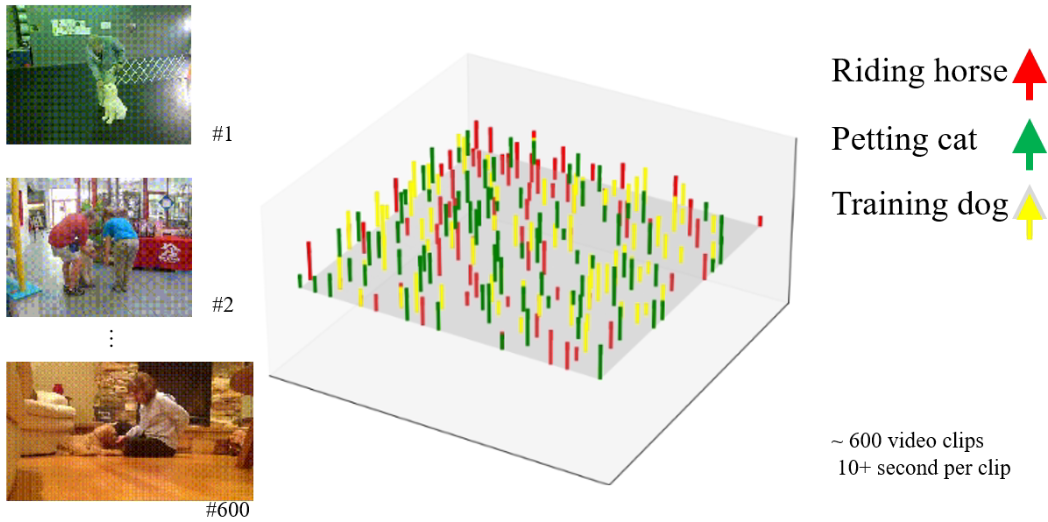}
    \caption{Training dog.}
    \label{fig:training_dog}
\end{figure}

\begin{figure}
    \centering
    \includegraphics[width=0.575\linewidth]{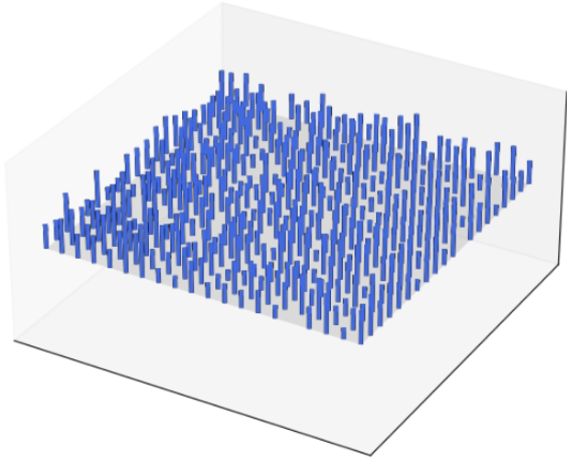}
    \caption{Universal Action Space.}
    \label{fig:universal_action_space}
\end{figure}

\begin{figure}
    \centering
    \includegraphics[width=1\linewidth]{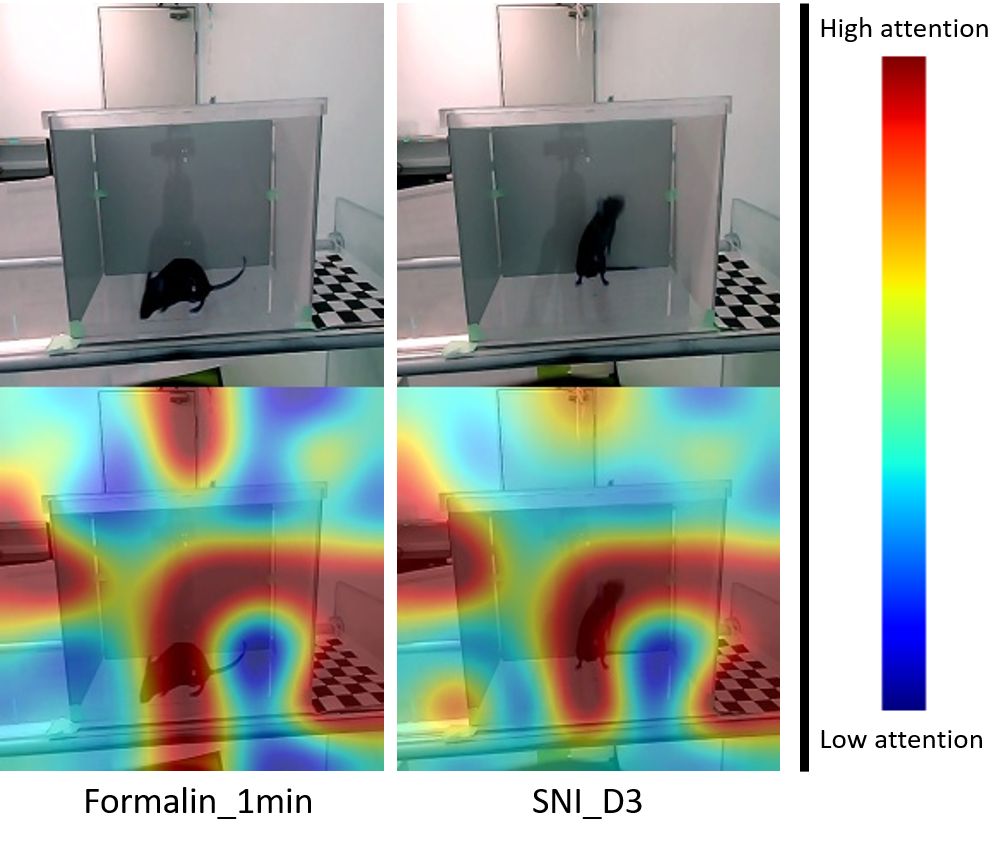}
    \caption{Action attention maps with colormap.}
    \label{fig:attention_maps_with_colormap}
\end{figure}

\begin{figure}
    \centering
    \includegraphics[width=1\linewidth]{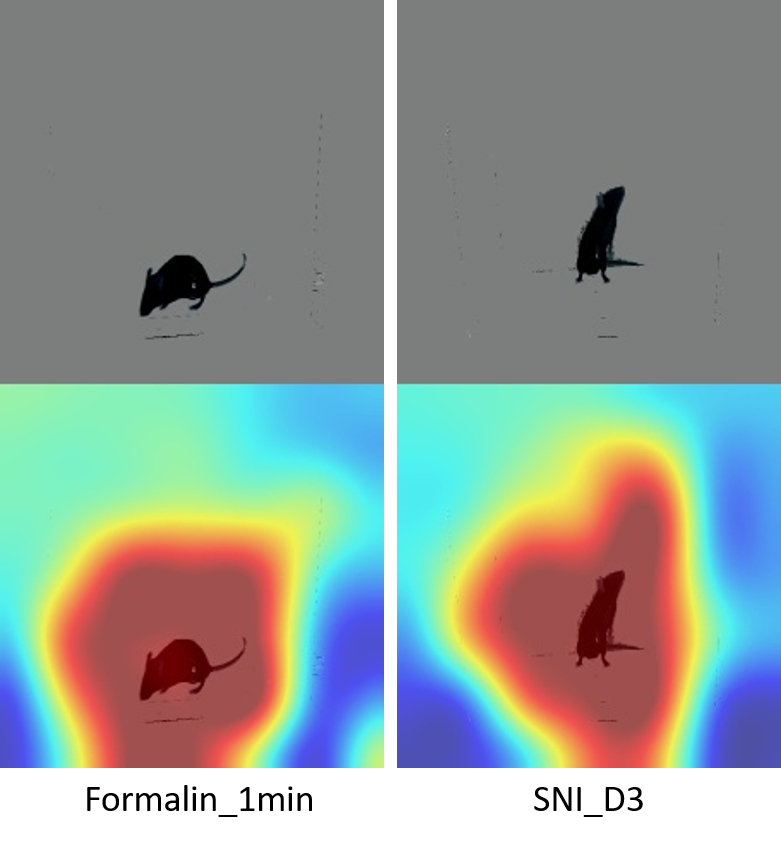}
    \caption{Pure foreground (mouse) filtered by temporal median filtering. Bottom row are the corresponding action attention maps detected by VST.}
    \label{fig:foreground_mouse}
\end{figure}

\begin{figure}
    \centering
    \includegraphics[width=1\linewidth]{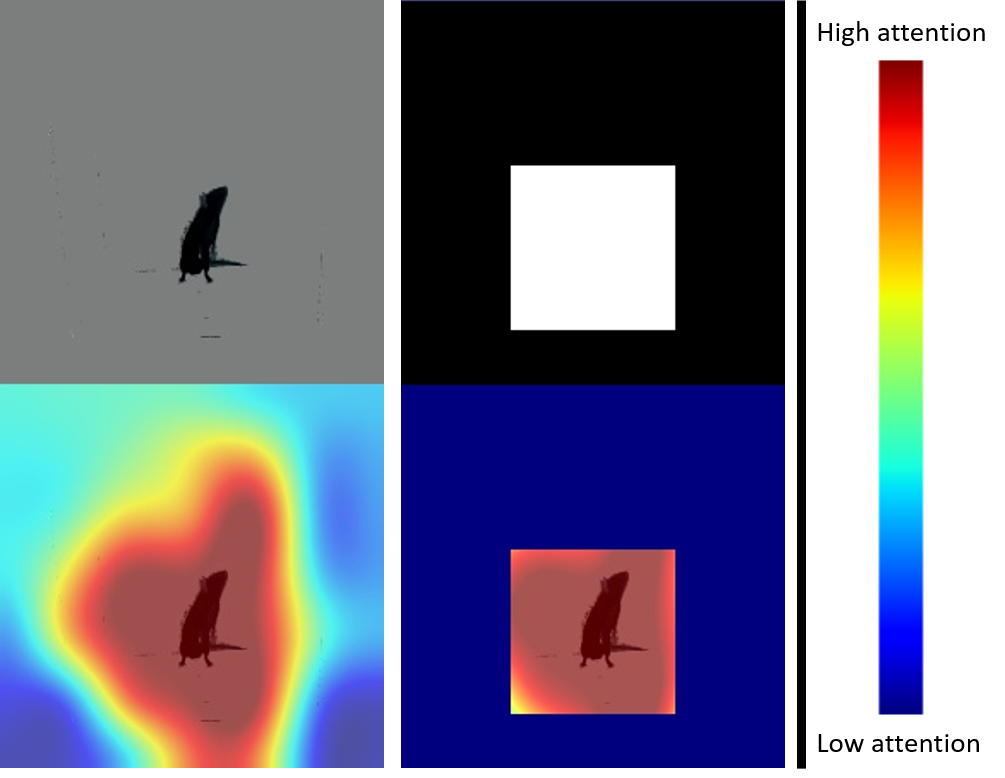}
    \caption{
    Top-left: foreground mouse image; 
    Bottom-left: action attention map from VST over the foreground image; 
    Top-right: 0/1 spatial mask applied to VST features; 
    Bottom-right: attention map after applying the mask on VST outputs.
    }\label{fig:masking_features_with_attention}
\end{figure}

In this section, 
we will introduce how to construct UAS using Kinetics-600, 
a video dataset of complex human behaviors. 
Kinetics-600 records 600 
types of human behaviors, 
such as riding horses, 
playing tennis, 
training dog, 
and so on. 
Each member of Kinetics-600 consists of about 600 video clips of 5-10 seconds.
We use VST~\cite{video_swin_transformer_liu2021videoswintransformer} to extract spatial-temporal features from the training videos and project them onto UAS. 
For example, Figure \ref{fig:riding_horse} shows a total of 600 video clips, each about 10 seconds long, labeled as horse riding.
These video clips were trained one by one and incrementally projected onto the UAS they are to construct. 
Similarly, we also trained and projected 600 petting cats and 600 training dog video clips to the UAS in sequence, 
and the corresponding UAS incremental construction process is shown in Figure \ref{fig:petting_cat} and Figure \ref{fig:training_dog} respectively. After all 600 human behavior video clips of Kinetics-600 are used for training, the appearance of UAS is shown in Figure \ref{fig:universal_action_space} (schematic diagram). 
We can imagine that the UAS constructed in this way is a very large action space, 
and we can boldly assume that it can include the action space constructed by the simpler actions of most other animals.
To be more specific, the spaces constructed by the simpler actions of other animals are all its subspaces.

\subsection{Construction of Mice Action Space on UAS}
\label{Construction_of_Mice Action_Space_on_UAS}
The methodology developed in this work can be applied to all animal behavioral analysis.
Here we use the analysis of chronic pain in mice as an example.
As described in Section \ref{sec:data_collection_settings},
we collected a total of 632 video clips,
and they belong to the mice with healthy condition, 
neuralgia pain,
and inflammatory pain respectively (Table \ref{tab:dataset}).
As for the 632 five-minute videos,
we cut these videos into equal parts according to the number of each category,
and later we shall perform 8-fold cross validation~\cite{k_fold_WONG20152839}.
In the experiments,
we randomly select 6 parts as training data,
one as validation data,
and one as testing data.
During the training phase,
we used the UAS constructed in Section \ref{subsec:construction of UAS} as a basis and divided each five-minute mouse training video into two hundreds 1.5-second video segments,
to train a system that can recognize chronic pain in mice.

Because we use VST to capture spatio-temporal features in videos to train the mouse action space,
it is very important that the mouse action content it captures is correct. 
Figure \ref{fig:attention_maps_with_colormap} shows the action attention maps captured by the VST using Grad-CAM \cite{grad_cam_selvaraju2020grad} at two different time instances, 
and the color of the action attention maps represents the intensity of the action.
The redder the area,
the more significant the action. 
In the second row of Figure \ref{fig:attention_maps_with_colormap},
most of the areas where the mouse is located can be defined as areas with actions by VST. However,
VST also mislabels some background areas. In order to use more accurate training data to train the mouse action subspace on the UAS,
we first execute a temporal median filtering process.
For each five-minute long mouse video clip,
% which consists of 6000 frames,
we take the median of the brightness values of every pixel (out of all frames in the 5-minute video),
making a pure five-minute background video.
For each five-minute mouse training video,
we use the above background video to take their difference, 
the remaining difference video will be the pure mouse actions,
as shown in the first row of Figure \ref{fig:foreground_mouse}.
The second row of Figure~\ref{fig:foreground_mouse} shows the action attention maps corresponding to the mouse instances of the first row of Figure~\ref{fig:foreground_mouse}.
Obviously, the original troubles caused by the background have been greatly improved.

However, as can be observed from the lower half of Figure~\ref{fig:foreground_mouse}, when VST is applied to the mouse video with the background removed, 
there is still a large area around the target object (mouse) where non-negligible motion is detected.
To help VST focus more on the mouse during operation, we introduce a feature-space foreground mask. This mask is defined in VST's feature space and serves to guide the system's attention. This mask will find the $3\times3$ block that have the greatest response to the mask from the larger action area (redder areas in the bottom of Figure~\ref{fig:foreground_mouse}) that was originally divided into $7\times7$ blocks. As can be seen from the upper right and lower right corners of Figure~\ref{fig:masking_features_with_attention}, the above mentioned $3\times3$ block is indeed centered on the mouse, significantly reducing the original impact range. Because this range is reduced, the impact of the mouse's motions will be greatly increased, which also means that the movements that VST can capture will be more accurate.

We use the VST to capture the spatio-temporal information from the above processed video with only mouse actions and construct the mouse's action subspace on UAS.
Then, we used the videos in the validation set to check whether the constructed mouse action space on UAS is appropriate or not.

\begin{figure}
    \centering
    \includegraphics[width=1\linewidth]{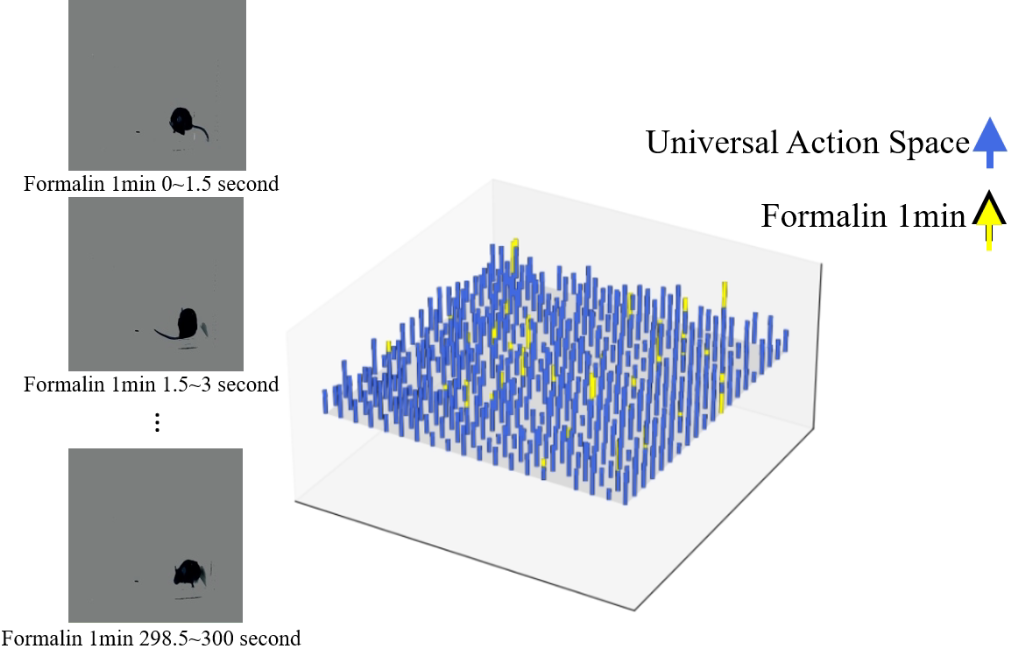}
    \caption{Formalin 1min action space.}
    \label{fig:formalin_1min_space}
\end{figure}

\begin{figure}
    \centering
    \includegraphics[width=1\linewidth]{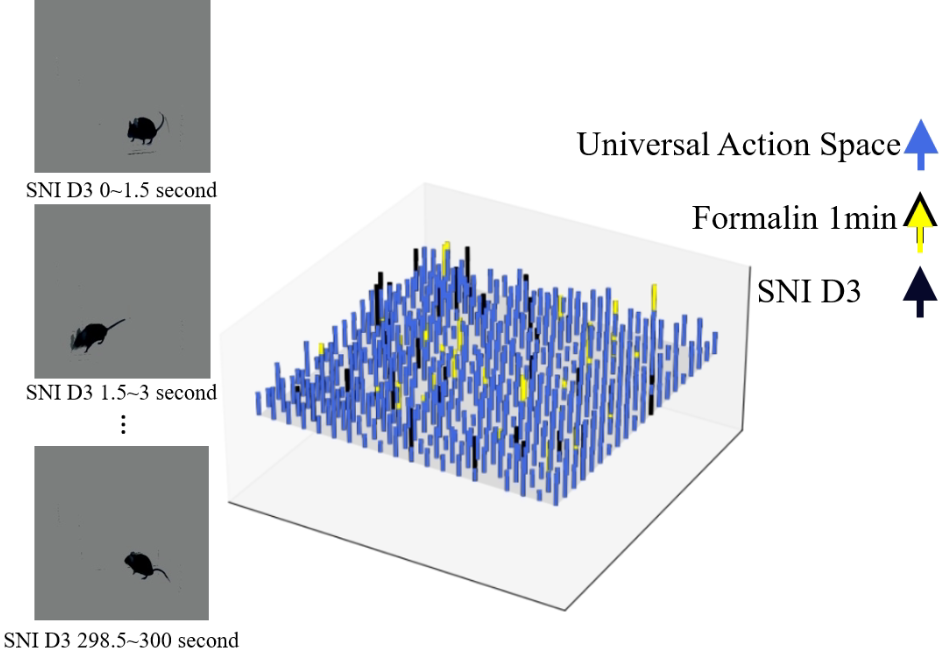}
    \caption{SNI D3 action space.}
    \label{fig:sni_d3_space}
\end{figure}

\begin{figure}
    \centering
    \includegraphics[width=1\linewidth]{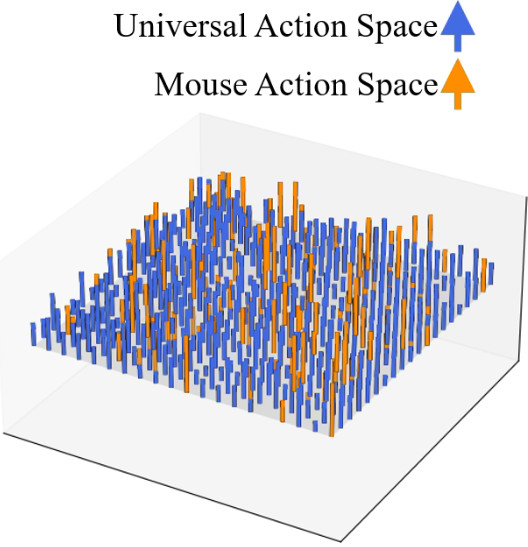}
    \caption{Universal Action Space with mouse action space.}
    \label{fig:universal_action_space_with_mouse_action_space}
\end{figure}

The way we construct the mouse's action subspace on UAS is as follows. 
Figure \ref{fig:formalin_1min_space} shows the subspace constructed using 56 five-minute videos (only use the training part) of mouse behavior (one-minute after formalin injection, schematic diagram).
Then we trained the mouse subspace using 60 videos (training part only) recorded from different mice (three-day after SNI injection),
the corresponding incremental subspace is shown in Figure \ref{fig:sni_d3_space} (schematic diagram).
After all 470 five-minute-long training videos are trained,
the possible appearance of the subspace constructed on UAS is shown in Figure \ref{fig:universal_action_space_with_mouse_action_space} (schematic diagram).

%% file: sec/4_method_mm_eng.tex
\section{Experiments}
\label{sec:Experiments}

\subsection{Evaluation Metrics}

We use Micro Accuracy and Macro F1-score~\cite{macro_f1_opitz2021macrof1macrof1} to evaluate the performance of the system. 
Micro accuracy shows the accuracy of the overall mouse behavior identification, while Macro F1-score shows the cause of each pain and the pain analysis at different lengths of time.  
We statistically analyze the system's performance in identifying the cause of pain and the duration of pain.  
The causes of pain include no pain, inflammatory pain, and neuralgic pain.  
The pain duration identification further estimates the identification of three causes of pain sampled at 15 time points to conduct a more detailed analysis of acute and chronic pain.

\subsection{Main Results}
Our test set is the 1/8 data split out from the above mentioned 8-fold cross validation~\cite{k_fold_WONG20152839} specifically for testing.
To prevent VST from accidentally capturing incorrect action information from the background (Figure \ref{fig:attention_maps_with_colormap}),
we performed an additional step similar to the training data,
i.e., performing temporal median filtering on the five-minute each testing video. 
At the same time, we also use feature-space masks to further suppress those background motions that were mistakenly detected.
After these steps, it is equivalent to extracting spatio-temporal features directly from the foreground video that contains only mouse movements,
and then performing classification on the mouse action subspace built on the UAS.

In the experiments, we will mark the method of removing background by using temporal median filtering as "Ours", and if we further add the feature space masking step, we mark it as "Ours+Mask".
We compare the proposed method with the state-of-the-art method (B-SOiD~\cite{b_soid_Hsu_2021}) and the result of human expert interpretation.
The human expert raters were trained members of the biology laboratory research team (15 members in total) who had experience in mouse behavior analysis but were not trained as expert annotators.
The expert raters randomly selected 10 videos from the test video dataset (each with equal probability) through a process,
and then determined which of the 15 categories they belonged to.
It is very difficult to make such a detailed classification of healthy and pain mice. In terms of systematic methods, we used the same set of mouse videos to train the state-of-the-art method B-SOiD~\cite{b_soid_Hsu_2021}, which employs DeepLabCut~\cite{deep_lab_cut_mathis2018deeplabcut} to track 12 anatomical landmarks on the mouse body, and paired it with ViS4mer~\cite{vis4mer_islam2023longmovieclipclassification} for mouse pain analysis.

\begin{table}[h]
    \centering
    \resizebox{0.48\textwidth}{!}{
    \begin{tabular}{|l|c|c|c|c|}
        \hline
        \textbf{Method} &  \textbf{15-class Acc} & \textbf{15-class F1} & \textbf{3-class Acc} & \textbf{3-class F1} \\
        \hline
        Human Expert  & 21.33 & - & 48.00 & -  \\
        \hline
        B-SOiD~\cite{b_soid_Hsu_2021}   & 30.52 & 0.1111 & 58.43 & 0.5645 \\
        \hline
        Ours & 43.03 & 0.3275 & 68.03 & 0.6706 \\
        Ours+Mask & 48.41 & 0.3863 & 73.10 & 0.7144 \\
        \hline
    \end{tabular}
    }
    \caption{Comparison}
    \label{table:main_ex}
\end{table}

Because even human experts or B-SOiD method find it difficult to achieve satisfactory results in classifying pain into 15 categories, 
we conducted another set of experiments.
In this set of experiments,
the pain classification was changed from the original 15 categories to just 3 categories: inflammatory pain, neuropathic pain, and no pain.
We show the experimental results in Table \ref{table:main_ex}.

When the classification is divided into 15 categories, the human expert, B-SOiD + ViS4mer, Ours, and Ours+Mask methods achieved 21.33\%, 30.52\%, 43.03\% and 48.41\% accuracy in Micro Accuracy, respectively.
As for the performance on Macro F1-score, B-SOiD + ViS4mer, Ours, and Ours+Mask methods achieved 0.1111, 0.3275, and 0.3863 respectively. 
In the classification of 15 categories, our methods have a significant advantage in both Micro Accuracy and Macro F1-Score.
In addition, in the experiment with only three categories, human expert, B-SOiD+ViS4mer, Ours, and Ours+Mask methods achieved classification rates of 48.00\%, 58.43\%, 68.03\%, and 73.10\% in the Micro Accuracy experiment, respectively.
On the three-class classification using Macro F1-score as the evaluation criterion, B-SOiD+ViS4mer, Ours, and Ours+Mask methods obtain 0.5645, 0.6706, and 0.7144 respectively.
From the above analysis, our method has obvious advantages over B-SOiD + ViS4mer in both Micro Accuracy and Macro F1-score calculation processes.
We also found that the VST would extract incorrect motion information from the background. This unnecessary information would dilute the influence of the correct information during training and classification, resulting in inaccurate results.

\subsection{Zero-shot Evaluation on Pain Medication}
\label{subsection:pain_medication}

\begin{figure}
    \centering
    \includegraphics[width=0.6\linewidth]{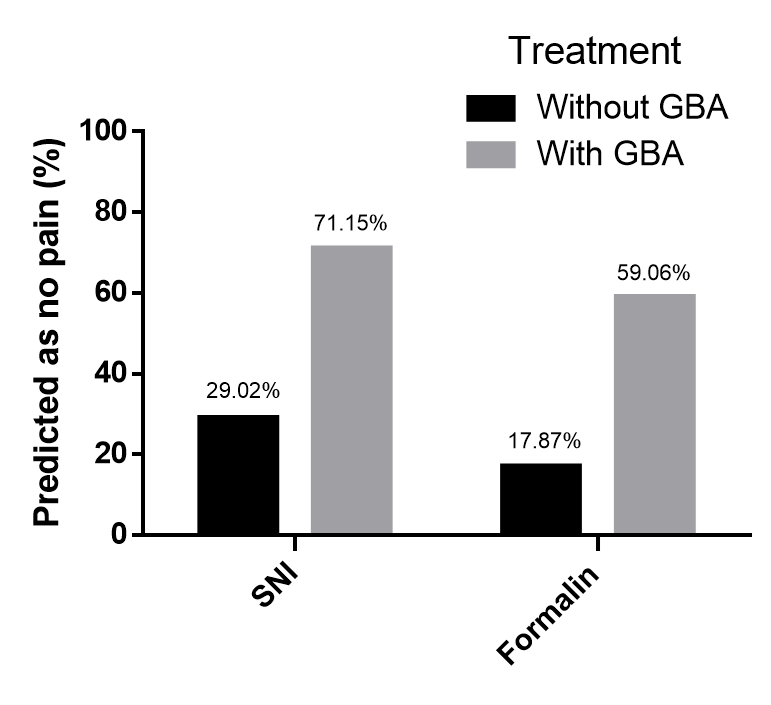}
    \caption{Visualization of pain in mice with and without medicine.}
    \label{fig:medicine_bf_aft}
\end{figure}

\begin{figure}
    \centering
    \includegraphics[width=1.0\linewidth]{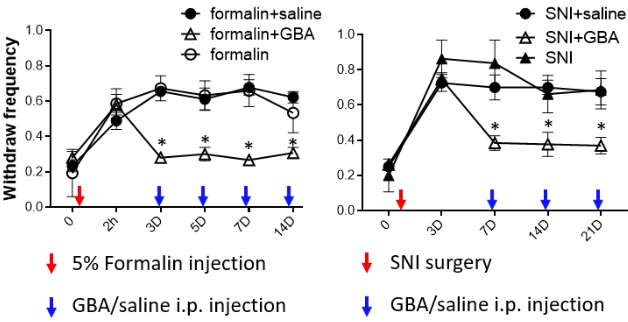}
    \caption{Von Frey mechanical pain test results for mice treated with Gabapentin (GBA) and Saline following formalin and SNI procedures. (* p $<$ 0.05 between GBA and Saline groups.)}
    \label{fig:GBA_von_frey}
\end{figure}

To evaluate the potential application of our proposed mouse pain behavior analysis model in drug testing,
we conducted a zero-shot experiment,
applying the pre-trained model to assess behavioral responses following analgesic drug administration.
In this experiment,
mice with neuropathic pain (n = 13) and inflammatory pain (n = 10) were  administered Gabapentin (GBA, i.p. injection 50mg/kg),
a drug widely used for the treatment of neuropathic pain~\cite{gba_for_sni_sadler2019gabapentin,gba_for_sni_scuteri2021effect},
or normal saline (n = 8) as a control.
Five-minute behavioral videos were recorded 30 minutes post-injection and analyzed using the previously trained model.
As shown in Figure~\ref{fig:medicine_bf_aft},
59.06\% of mice in the formalin group were classified as “no pain” 30 minutes after GBA injection,
significantly higher than the 17.87\% observed in untreated formalin day-3 mice.
Similarly, 71.15\% of GBA-treated SNI mice were classified as “no pain,” compared to only 29.02\% in untreated SNI day-7 mice.
These findings are consistent with the results of Von Frey mechanical sensitivity tests (Figure~\ref{fig:GBA_von_frey}).

Further summarized in Table~\ref{table:pain_medication}, model predictions across different pain types revealed that approximately 30\% of saline-injected mice 
exhibited a classification shift, 
with some identified as “no pain.”
We speculate that this phenomenon may result from two factors:
(1) the restraint and injection procedure itself may influence mouse behavior;
(2) recent studies have suggested that saline injections may exert mild therapeutic effects in human chronic pain treatment~\cite{human_saline_bar2017use,huamn_saline_suputtitada2023intra,human_saline_gazendam2021intra,human_saline_kongsagul2020ultrasound}.
Consistently, Von Frey testing showed a slight reduction in mechanical pain sensitivity in the saline-treated neuropathic pain group (Figure~\ref{fig:GBA_von_frey}),
although no such effect was observed in the inflammatory pain group.

Together, these results suggest that our automated mouse pain classification model possesses strong application potential and may serve as an effective tool for evaluating the efficacy of analgesics in preclinical studies.

\begin{table}[h]
\centering
\caption{Classification Results on Mice Treated with Pain Medication}
\label{table:pain_medication}
\resizebox{0.48\textwidth}{!}{ % Adjust table size for two-column format
\begin{tabular}{|l|c|c|c|}
\hline
\textbf{Ground Truth (GT)} & \textbf{Formalin (\%)} & \textbf{Control (\%)} & \textbf{SNI (\%)} \\
\hline
    SNI & 11.11 & 34.64 & 54.25 \\
    SNI + Saline & 31.25 & 51.56 & 17.18 \\
    SNI + GBA & 24.35 & 71.15 & 4.48 \\
    % \hline
    Formalin & 63.40 & 30.80 & 5.80  \\
    Formalin + Saline & 37.50 & 56.25 & 6.25 \\
    Formalin + GBA & 38.28 & 59.06 & 2.65 \\

\hline
\end{tabular}
}
\end{table}

%% file: sec/6_ana_dis_mm_eng.tex
\section{Analysis and Discussion}
\label{sec:ana_dis}

\subsection{Behavioral Analysis of acute pain and chronic pain}
\label{sec:acute_pain_and_chronic_pain}

\begin{figure}
    \centering
    \includegraphics[width=1\linewidth]{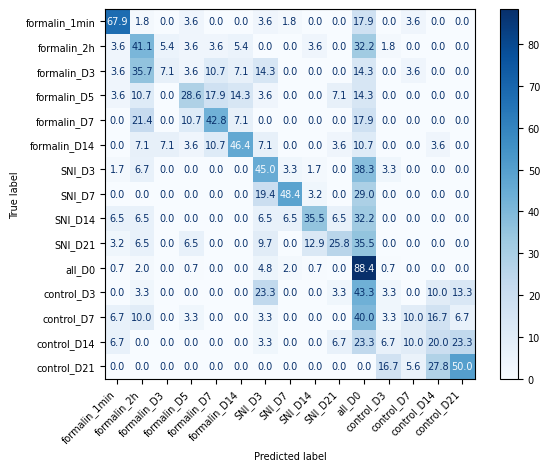}
    \caption{Normalized confusion matrix for 15-class mouse pain behavior classification.}
    \label{fig:conf_matrix_ours}
\end{figure}

We show the confusion matrix of mouse pain behavior classification in Figure~\ref{fig:conf_matrix_ours} to further evaluate the necessity of dividing behaviors into 15 categories and to gain insights that may guide future classification strategies.
The matrix also provides an opportunity to examine the similarities and differences between acute and chronic pain behaviors.
The results show that most misclassifications occurred within behavior groups derived from the same pain model and typically between adjacent time points.
This suggests that behaviors induced by the same pain source are highly similar,
making them harder to distinguish.
For example, in the control group,
we observed significant confusion among days 3, 7, 14, and 21, indicating minimal behavioral differences across these time points and implying that such detail categorization may not be necessary.
In the formalin model, classification accuracy was highest at 1 min post-injection,
reflecting the presence of pronounced paw-licking behavior during the acute pain phase—an easily detectable feature for the model. In contrast,
formalin days 5, 7, and 14 exhibited greater confusion,
suggesting these time points represent later phases of formalin-induced pain where movement features become less distinct and harder to classify.
In the SNI, we found that behaviors at days 14 and 21 were misclassified as healthy (control),
most likely because mouse movement patterns at these time points closely resembled those of non-painful conditions. Additionally,
we observed partial confusion between SNI mice and formalin mice.
This overlap may reflect shared similar behavioral features across different pain models and presents a compelling direction for future investigation.

\subsection{Analysis of Pain Progression in Individual Mice}

In~\ref{sec:acute_pain_and_chronic_pain}, we observed that the main reason for the misclassification of pain behavior in mice is the misclassification of the number of pain days, and we believe that this is caused by the differences in pain tolerance from different mice. 
Therefore, we further use Quadratic Weighted Kappa (QWK) to analyze each mouse's pain time growth and the temporal consistency with the model prediction results and show the analysis results of mice with different pain causes in Table~\ref{tab:mean_qwk_kappa}. 
We cut the time length into three groups, D0 to D3 (acute pain), D3 to D7, and D7 to D21 (chronic pain), and calculated QWK according to the mouse pain category. 
It can be seen from the results that although the original 15-class F1 score of our "Ours+Mask" method in Table~\ref{table:main_ex} is only 0.3863,
the consistency analysis of the overall pain level and the number of pain days can reach 0.4463. 
This shows that the proposed model can moderately
capture the changes in pain levels of each mouse over time. 
In addition, the QWK index was generally higher in acute pain than in chronic pain for the pain-inducing conditions (Formalin and SNI). 
For the control group, this pattern is not observed because the mice do not exhibit pain behavior at all, 
and therefore, there is no distinction between acute and chronic phases. 
These results suggest that chronic pain remains more challenging to capture accurately, 
and improving model performance in this phase is an important direction for future research.

\begin{table}[h]
    \centering
    \caption{Mean Quadratic Weighted Kappa Scores by Condition and Pain Phase}
    \label{tab:mean_qwk_kappa}
    \resizebox{0.48\textwidth}{!}{ % Adjust table size for two-column format
    \begin{tabular}{|l|c|c|c|c|}
    \hline
    Condition & Acute Pain (D0 to D3)& D3 to D7 & Chronic Pain(D7 to D21) & Overall \\
    \hline
    Formalin & 0.5099 & 0.3222 & 0.3730 & 0.5291 \\
    SNI & 0.4561 & 0.3978 & 0.3151 & 0.3365 \\
    Control & 0.1724 & 0.0925 & 0.3350 & 0.5097 \\
    Total & 0.4194 & 0.2711 & 0.3400 & 0.4463 \\
    \hline
    \end{tabular}
    }
\end{table}

%% file: sec/7_con_mm_eng.tex
\section{Conclusions}
\label{sec:con}

This study proposes a chronic pain behavior analysis framework based on deep learning, which provides a more objective and data-driven way to identify mouse pain through the combination of short-term action feature extraction and long-term behavior modeling.
Experimental results show that a large-scale action feature space contains richer semantics performs better in mouse pain recognition, which shows that an action space trained with a large amount of data can more effectively capture the subtle action features of mice.  The results show that this system even surpasses the classification capabilities of the most popular B-SOiD method and human experts, which also proves the extraordinary value of end-to-end deep learning for animal behavior analysis. 

Our model also successfully demonstrated robust generalization ability across unseen drug intervention profiles, accurately capturing pain-relieving behavioral patterns especially in the GBA group, which also shows its potential in drug efficacy evaluation.  In the future, we plan to further expand the dataset to different animals and pain models to improve generalization capabilities.  This research method is not only suitable for pain behavior analysis, but also provides a powerful and interpretable analysis tool for other animal behavior studies.